\documentclass[letterpaper]{article} % DO NOT CHANGE THIS
\usepackage{aaai25}  % DO NOT CHANGE THIS
\usepackage{times}  % DO NOT CHANGE THIS
\usepackage{helvet}  % DO NOT CHANGE THIS
\usepackage{courier}  % DO NOT CHANGE THIS
\usepackage[hyphens]{url}  % DO NOT CHANGE THIS
\usepackage{graphicx} % DO NOT CHANGE THIS
\usepackage{amsmath}
\usepackage{amssymb} % 包含 \mathbb 命令的包
\usepackage{amsfonts} % 另一个包含 \mathbb 命令的包
\usepackage[ruled,linesnumbered]{algorithm2e}
\usepackage{natbib}  % DO NOT CHANGE THIS AND DO NOT ADD ANY OPTIONS TO IT
\usepackage{caption} % DO NOT CHANGE THIS AND DO NOT ADD ANY OPTIONS TO IT
\usepackage{amsmath}

\usepackage{newfloat}
\usepackage{listings}

\newtheorem{theorem}{Theorem}
\DeclareCaptionStyle{ruled}{labelfont=normalfont,labelsep=colon,strut=off} % DO NOT CHANGE THIS
\title{TRAIL: Trust-Aware  Client Scheduling for  Semi-Decentralized Federated Learning  }
\author{
Gangqiang Hu\textsuperscript{\rm 1}, Jianfeng Lu\textsuperscript{\rm 1,2,3}\thanks{Corresponding Authors are Jianfeng Lu and Jianmin Han. \\   The full version is at https://arxiv.org/abs/2412.11448.},  Jianmin Han\textsuperscript{\rm 1}$^{*}$, Shuqin Cao\textsuperscript{\rm 2}, Jing Liu\textsuperscript{\rm 4}, Hao Fu\textsuperscript{\rm 4} 
}
\affiliations{
\textsuperscript{\rm 1}School of Computer Science and Technology, Zhejiang Normal University, China\\
\textsuperscript{\rm 2}School of Computer Science and Technology, Wuhan University of Science and Technology, China\\
\textsuperscript{\rm 3}Key Laboratory of Social Computing and Cognitive Intelligence (Dalian University of Technology), Ministry of Education, China\\
\textsuperscript{\rm 4}Hubei Province Key Laboratory of Intelligent Information Processing and Real-time Industrial 
System (Wuhan Universityof Science and Technology), China\\
\{hugangqiang, lujianfeng, hanjm\}@zjnu.edu.cn, \{shuqincao, luijing\_cs,fuhao\}@wust.edu.cn
}

\usepackage{bibentry}
\begin{document}

\maketitle

\begin{abstract}

Due to the sensitivity of data, Federated Learning (FL) is employed to enable distributed machine learning while safeguarding data privacy and accommodating the requirements of various devices. 
However, in the context of semi-decentralized FL, clients' communication and training states are dynamic. This variability arises from local training fluctuations, heterogeneous data distributions, and intermittent client participation. Most existing studies primarily focus on stable client states, neglecting the dynamic challenges inherent in real-world scenarios. 
To tackle this issue, we propose a \underline{TR}ust-\underline{A}ware cl\underline{I}ent schedu\underline{L}ing mechanism called TRAIL, which assesses client states and contributions, enhancing model training efficiency through selective client participation. 
We focus on a semi-decentralized FL framework where edge servers and clients train a shared global model using unreliable intra-cluster model aggregation and inter-cluster model consensus. 
First, we propose an adaptive hidden semi-Markov model to estimate clients' communication states and contributions. Next, we address a client-server association optimization problem to minimize global training loss. Using convergence analysis, we propose a greedy client scheduling algorithm. 
Finally, our experiments conducted on real-world datasets demonstrate that TRAIL outperforms state-of-the-art baselines, achieving an improvement of 8.7\% in test accuracy and a reduction of 15.3\% in training loss.
\end{abstract}

% Uncomment the following to link to your code, datasets, an extended version or similar.
%
% \begin{links}
	%     \link{Code}{https://aaai.org/example/code}
	%     \link{Datasets}{https://aaai.org/example/datasets}
	%     \link{Extended version}{https://aaai.org/example/extended-version}
	% \end{links}

\section{Introduction}

The integration of advanced communication technologies with industrial manufacturing significantly enhances production efficiency and flexibility, accelerating the transition to smart manufacturing \cite{chen2024feddat,lu2021toward,tan2023federated}. This integration facilitates seamless connectivity between devices and systems through real-time data collection and analysis, which greatly improves the transparency and controllability of production processes \cite{yu2023clustered,wang2020optimizing}. Additionally, the incorporation of Artificial Intelligence (AI) further enhances these capabilities. By enabling systems to process and analyze large volumes of data, AI provides solutions for predictive maintenance, intelligent decision-making, and process optimization \cite{wu2024virtual}.

In modern AI systems, local data on end devices often contains sensitive or private information, rendering traditional edge AI training architectures impractical \cite{zhang2024vertical,wang2024feddse}. To address security and privacy concerns while minimizing communication costs, a new distributed machine learning framework called Federated Learning (FL) has emerged \cite{wu2023faster, McMahan2016CommunicationEfficientLO,wang2024fednlr}. In FL, each client uploads only model parameters, safeguarding her local data. Typically, this process involves coordination with a single edge server, which can result in high communication overhead and potential single points of failure, particularly in environments with numerous end devices \cite{zhang2023delving,lu2022toward}.

%In modern AI systems, local data on end devices often contains sensitive or private information, making traditional edge AI training architectures impractical \cite{zhang2024vertical,wu2023faster}. To address security and privacy issues while reducing communication costs, a new distributed machine learning framework known as SD-FL (FL) has been introduced \cite{wu2023faster, McMahan2016CommunicationEfficientLO}. In FL, each client uploads only model parameters, without exposing their local data. Typically, this process involves coordination with a single edge server, which can lead to high communication overhead and potential single points of failure, especially in environments with numerous end devices \cite{zhang2023delving,lu2022toward}.

%In the context of IIOT, local data on end devices often contains sensitive or private information, rendering traditional edge AI training architectures impractical \cite{zhang2024vertical,wu2023faster}. To tackle security and privacy issues while cutting communication costs, a new distributed machine learning framework called SD-FL (FL) has been introduced in IIoT \cite{wu2023faster, McMahan2016CommunicationEfficientLO}. In FL, each client only needs to upload model parameters without exposing their local data. Typically, this process involves coordination with a single-edge server. This coordination can lead to high communication overhead and potential single points of failure, especially in IIoT environments with many end devices \cite{zhang2023delving,lu2022toward}.

This paper investigates semi-decentralized FL (SD-FL) as a framework to enhance the reliability of model training. As illustrated in Figure \ref{sys}, we focus on a multi-edge server and multi-client SD-FL framework \cite{Sun2021SemiDecentralizedFE}. 
This framework utilizes a two-tier aggregation approach. The first tier is intra-cluster aggregation, where local models are aggregated by their respective servers. The second tier is inter-cluster consensus, which involves exchanging models from multiple servers and collaboratively aggregating them to train a shared global model. By distributing computational and communication loads, this approach enhances both the robustness and scalability of the FL process.
%This architecture employs a two-tier aggregation approach: first, intra-cluster aggregation, where local models are aggregated by their respective servers, followed by inter-cluster consensus, where models from multiple servers are exchanged and collaboratively aggregated to train a shared global model. By distributing computational and communication loads, SD-SD-FL improves both the robustness and scalability of the FL process.        
While SD-FL mitigates risks associated with single points of failure, existing research often overlooks the dynamic nature of clients, particularly fluctuations in model contributions and communication quality, which can adversely affect training efficiency \cite{sun2023semi}. 

Research was conducted to address the issue of unreliable clients. In \cite{Sefati2021AQS}, the authors introduced an effective service composition mechanism based on a hidden Markov model (HMM) and ant colony optimization to tackle IoT service composition challenges related to Quality-of-Service parameters. This approach achieved significant improvements in availability, response time, cost, reliability, and energy consumption. In \cite{Ma2021FedSAAS}, the FedClamp algorithm was proposed, which enhanced the performance of the global model in FL environments by utilizing HMM to identify and isolate anomalous nodes. This algorithm was specifically tested on short-term energy forecasting problems. The authors of \cite{Vono2021QLSDQL} presented the Quantized Langevin Stochastic Dynamics (QLSD) algorithm, which employed Markov Chain Monte Carlo methods to improve dynamic prediction capabilities in FL while addressing challenges related to privacy, communication overhead, and statistical heterogeneity. Additionally, \citeauthor{Wang2024TrustAoIAwareCO} introduced a trust-Age of Information (AoI) aware joint design scheme (TACS) aimed at enhancing control performance and reliability in wireless communication networks within edge-enabled Industrial Internet of Things (IIoT) systems operating in harsh environments. This scheme utilized a learning-based trust model and scheduling strategy. While these studies explored various dynamic aspects, they did not adequately address the interplay between dynamics and client selection strategies.

To bridge this gap, we propose an adaptive hidden semi-Markov model (AHSMM) to predict dynamic changes in training quality and communication quality. AHSMM enhances standard HMM by explicitly modeling state duration distributions, reducing computational complexity, and adapting to dynamic, multi-parameter environments, making it ideal for complex scenarios. To improve SD-FL systems' control and reliability, we propose a joint mechanism combining dynamic prediction with client selection. Extensive experiments and analyses validate the effectiveness and robustness of our approach under varying client dynamics. The main contributions of this paper are as follows:

%To address the gap above, we utilize a Markov model to predict the dynamic changes in client performance and communication quality and schedule clients based on the predicted results to minimize loss. We propose a joint design scheme combining dynamic prediction with client selection, significantly enhancing FL systems' control performance and reliability. Extensive experiments and theoretical analysis demonstrate the effectiveness and robustness of our mechanism under varying client dynamic conditions. The main contributions of this paper can be summarized as follows:

\begin{itemize}
%	\item We propose a unified optimization framework based on performance prediction and client scheduling, enhancing model robustness and convergence speed and improving overall model performance.
%	
%	\item We propose an adaptive hidden semi-Markov model (AHSMM) to predict client performance and channel variations. This model simultaneously considers both clients' dynamic and static aspects, enabling efficient state prediction for each client.
%	
%	\item By analyzing convergence, we determine the anticipated effects of client-server relationships on the convergence. From this analysis, we transform the initial optimization challenge into an  INLP problem and propose a greedy algorithm to optimize client scheduling efficiently.
%	\item  Our experiments on real-world datasets demonstrate the superiority of our proposed mechanism over the state-of-the-art baselines, achieving improvements of 8.7\% test accuracy and reducing 15.3\% training loss.       

\item We propose a unified optimization mechanism named TRAIL for SD-FL that integrates performance prediction and client scheduling, enhancing model robustness, accelerating convergence speed, and improving overall performance.

\item We introduce an  AHSMM to predict client performance and channel variations to obtain trust levels. This model effectively accounts for both dynamic and static aspects of clients, enabling efficient state predictions for each one.

\item Through convergence analysis, we assess the anticipated effects of client-server relationships on convergence. This analysis allows us to reformulate the initial optimization challenge as an integer nonlinear programming problem, for which we devise a greedy algorithm to optimize client scheduling efficiently.

\item Extensive experiments conducted on four real-world datasets (MNIST, EMNIST, CIFAR10, SVHN) demonstrate that our proposed mechanism outperforms state-of-the-art baselines, achieving an 8.7\% increase in test accuracy and a 15.3\% reduction in training loss.
\end{itemize}

\section{Related Work}

In FL, model training is distributed across multiple clients to protect data privacy and minimize the need for centralized data aggregation. Traditional FL assumed reliable and frequent communication between clients and the server. However, this assumption often failed in real-world applications, particularly in environments with heterogeneous devices and unstable communication. To address these challenges, researchers introduced SD-FL. This framework combined the benefits of centralized and distributed architectures by enabling direct communication among some clients, thereby reducing the server's workload and communication costs. SD-FL was better equipped to adapt to dynamic network environments and heterogeneous data distributions, enhancing the system's robustness and efficiency.

Research efforts primarily focused on the following two areas. (i) Client Selection: Mechanisms were developed to select clients for participation in training, ensuring that chosen clients met performance criteria, thereby enhancing the overall effectiveness of the model \cite{lin2021semi,Wang2024TrustAoIAwareCO,yemini2022semi,sun2023semi}. (ii) Trust Management: Trust mechanisms were introduced to assess and predict the reliability of clients, ensuring that only reliable clients participated in training. These mechanisms contributed to improved model robustness and performance \cite{beltran2023decentralized,parasnis2023connectivity,xu2024edge,valdeira2023multi}.

While some studies focus separately on client selection and trust management, existing methods often fail to effectively integrate these aspects, which negatively impacts the efficiency and performance of SD-FL systems. Our work proposes an integrated mechanism that addresses client selection and trust management, which is essential for enhancing the robustness and performance of SD-FL, especially in diverse and potentially unreliable environments.

\section{System Model}

\begin{figure}
	\centering
	\includegraphics[width=1\linewidth]{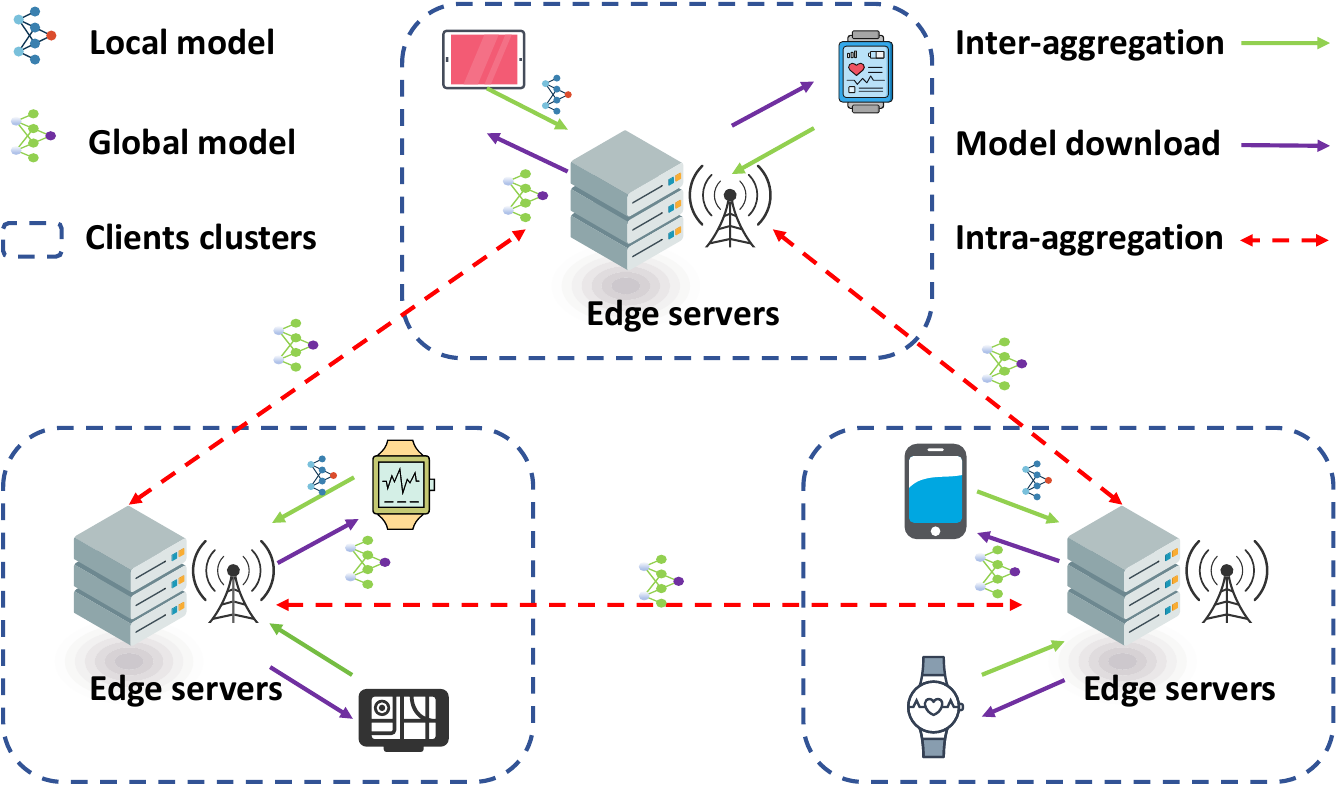}
	\caption{The SD-FL system framework.}
	\label{sys}
\end{figure}

Here, we explore the SD-FL framework, as illustrated in Figure \ref{sys}.   We first present SD-FL's basic workflow, then establish an adaptive semi-Markov model to estimate each client's model quality and communication quality.

% In the sequel, we elaborate on the preliminaries of SD-FL framework and system model respectively, and then formulate the optimization problem.

\subsection{Basics of SD-FL}

 We examine the SD-FL training process across $T$ rounds, involving $S$ edge servers represented by $\mathcal{S} = \{1, 2, \cdots, S\}$, and $U$ client devices represented by $\mathcal{U} = \{1, 2, \cdots, U\}$. Each round consists of the following steps:

\begin{itemize}
	\item Clients perform $T_1$ rounds of local training using their datasets. Then, they upload the trained local models to the edge server for intra-cluster aggregation. 
	\item After aggregating the models at the edge server, the merged model is broadcasted to the corresponding clients for model updating.

	\item After $T_2$ rounds of intra-cluster aggregation, each edge server sends its latest model to neighboring servers to achieve inter-cluster consensus.
	\item 
	After aggregating to obtain inter-cluster models, these models are sent back to their respective clients for the next round of training.
	
%	 According to the study \cite{Sun2021SemiDecentralizedFE}, it has been found that setting $T_1=T_2=1$ can reduce the number of training rounds required for convergence.

\end{itemize}

\subsection{AHSMM Model}

The Adaptive Hidden Semi-Markov Model (AHSMM) extends the traditional Hidden Semi-Markov Model (HSMM) \cite{McDonald2023ARN} into adaptive training using multi-parameter information (i.e.,  client training accuracy,  packet loss), thereby enhancing both modeling and analytical capabilities. The AHSMM model can be described by the parameters  $\Sigma = (\pi, A, B, E)$, where:
$\pi$ represents the initial state probabilities,
$A$ denotes the macro state transition probabilities,
$B$ corresponds to the observation probabilities after adaptive training,
$E$ represents the state dwell time after adaptive training, encompassing both the existing and remaining dwell times.
In addition, similar to HSMM, AHSMM addresses three core problems: evaluation, recognition, and training. To this end, AHSMM defines new forward-backward variables and proposes improved algorithms for forward-backward processes, the Viterbi algorithm \cite{zhang2023intelligent}, and the Baum-Welch algorithm \cite{zhang2023tensor}.

The computational complexity of the Hidden Semi-Markov Model (HSMM) is relatively high. To address this complexity, the Adaptive Hidden Semi-Markov Model (AHSMM) introduces a new forward variable, denoted as $\alpha_t(i, e)$. This variable represents the probability of generating the observations $z_1, z_2, \ldots, z_t$, given that the quality state $i$ has a specific dwell time of $e_t(i, e) = e$. In this context, $\varepsilon_t$ signifies the current dwell time of the quality state $q_t$.
%The computational complexity of the Hidden Semi-Markov Model (HSMM) is relatively high. To reduce this complexity, the Adaptive Hidden Semi-Markov Model (AHSMM) defines a new forward variable $\alpha_t(i, e)$, which represents the probability of generating the observations $z_1, z_2, \cdots, z_t$ given that the quality state $i$ has an existing dwell time $e_t(i, e) = e$. Here, $\varepsilon_t$ denotes the current dwell time of the quality state $q_t$.
%When $\left(q_t, \varepsilon_t\right)$ is assigned the value $(i, e)$, it signifies that the device maintains its current quality state $i$ up to time $t$, where the state $i$ has an accrued dwell time $e$, and is poised to shift to a different quality state at time $t+1$. Thus, for $1 \leq t \leq T-1$ and $e \in [1, E]$, we can define the forward variable as follows:
 When $\left(q_t, \varepsilon_t\right)$ is assigned the value $(i, e)$, it indicates that the device has remained in its current quality state $i$ up to time $t$. During this period, the state $i$ has accumulated a dwell time of $e$ and is prepared to transition to a different quality state at time $t+$ 1. Therefore, for $1 \leq t \leq T-1$ and $e \in[1, E]$, we can define the forward variable as follows:
 \begin{equation}
 	\alpha_t(i, e)=p\left(z_1^t,\left(q_{[t-e+1, t]}=i, \varepsilon_t=e\right) \mid \Sigma\right).  
 \end{equation}
The forward recursion is obtained as:
\begin{equation}
	\alpha_t(i, e) = 
	\begin{cases}
		\sum_{j \neq i}^N a_{ji} b_i(z_t) \left(\sum_{\tau=1}^E \alpha_{t-1}(j, \tau) p_j(\tau)\right), & \\
		\hspace{15em} \text{if } e = 1, \\
		\alpha_{t-1}(i, e-1) \prod_{s=1}^e b_i(z_{t-s+1}), & \\
		\hspace{15em} \text{if } e > 1, \\
		\pi_i b_i(z_1) p_i(e), & \\
		\hspace{15em} \text{if } t = 1, \\
		0, & \\
		\hspace{15em} \text{if } \tau < 1.
	\end{cases}
\end{equation}
%\begin{equation}
%\begin{aligned}
%	\alpha_t(i, e) &= p\left(z_1, z_2, \cdots, z_t, q_{[t-e+1, t]} = i \mid \Sigma\right) \\
%	&= p\left(z_1^t, \left(q_t, \varepsilon_t\right) = (i, e)\right).
%\end{aligned}
%\end{equation}
%\begin{equation}
%	\begin{aligned}
	%		\alpha_t(i, e) &= p\left(z_1, z_2, \cdots, z_t, q_{[t-e+1, t]} = i \mid \Sigma\right) \\
	%		&= p\left(z_1^t, \left(q_t, \varepsilon_t\right) = (i, e)\right).
	%	\end{aligned}
%\end{equation}

%\begin{equation}
%	\begin{aligned}
	%		\alpha_t(i, e) &= p\left(z_1, z_2, \ldots, z_t \mid \Sigma, q_{[t-e+1, t]} = i\right) \
	%		&= p\left(z_1^t, , (q_t, \varepsilon_t) \in {(i, e)}\right).
	%	\end{aligned}
%\end{equation}
%\begin{equation}
%	\alpha_t(i, e)= \begin{cases}\sum_{j \neq i}^N a_{j i} b_i\left(z_t\right)\left(\sum_{\tau=1}^E \alpha_{t-1}(j, \tau) p_j(\tau)\right), & \text { if } e=1, \\ \alpha_{t-1}(i, e-1) \prod_{s=1}^e b_i\left(z_{t-s+1}\right), & \text { if } e>1 .\end{cases}
%\end{equation}
%
%%\begin{equation}
%%\alpha_t(i, e) = \begin{cases}
%%	\sum_{j=1, j \neq i}^N a_{j i} b_i(z_t) \left(\sum_{\tau=1}^E \alpha_{t-1}(j, \tau) p_j(\tau)\right), & \\ \hspace{15.5em} \text{if }  e = 1, \\
%%	\alpha_{t-1}(i, e-1) \prod_{s=1}^e b_i\left(z_{t-s+1}\right), &  \\ \hspace{15.5em} \text{if } e > 1. \\
%%\end{cases}
%%\end{equation}
%
%
%\begin{equation}
%	\alpha_1(i, e)=\pi_i b_i\left(z_1\right) p_i(e),\\
%	\alpha_\tau(i, e)=0, \tau<1. 
%\end{equation}

In the context of AHSMM, let $E$ represent the maximum state dwell time among all quality states. Given the model $\Sigma$, the probability of observing the sequence $Z$ is expressed as:
\begin{equation}
	p\left(z_1^T \mid \Sigma\right)=\sum_{(i, e)} \alpha_T(i, e). 
\end{equation}
%In the context of AHSMM, let $E$ denote the maximum state dwell time among all quality states. Given the model $\Sigma$, the probability of the observation sequence $Z$ is given by:
%\begin{equation}
%p\left(z_1^{T} \mid \Sigma\right) = \sum_{i=1}^N \sum_{e=1}^E \alpha_T(i, e).
%\end{equation}
The variable $\alpha_T(i, e)$ is defined as the joint probability of observing the sequence $z_1, z_2, \ldots, z_T$ while the system is in quality state $i$ over the dwell time interval from $T-e+1$ to $T$. Mathematically, it can be expressed as:
%\begin{equation}
%	\alpha_T(i, e) = p\left(q_{[T-e+1, T]}=i \mid z_1, z_2, \ldots, z_T, \Sigma\right) \\
%	\cdot \, p\left(z_1, z_2, \ldots, z_T \mid \Sigma\right).
%\end{equation}
\begin{equation}
	\begin{aligned}
		\alpha_T(i, e) &= p\left(q_{[T-e+1, T]}=i \mid z_1, z_2, \ldots, z_T, \Sigma\right) \\
		&\cdot \, p\left(z_1, z_2, \ldots, z_T \mid \Sigma\right).
	\end{aligned}
\end{equation}
%In this scenario, $\alpha_T(i, e)$ is defined as:
%\begin{equation}
%\begin{aligned}
%	\alpha_T(i, e) &= p\left(z_1, z_2, \cdots, z_T, q_{[T-e+1, T]} = i \mid \Sigma\right).
%\end{aligned}
%\end{equation}
For $1 \leq t \leq T-1$, $e \in [1, E]$, and $i, j \in S$, the backward variable can be defined as:
\begin{equation}
	\beta_t(i, e)=p\left(z_{(t+1): T} \mid q_{[t-e+1, t]}=i, \Sigma\right).   
\end{equation}
%\begin{equation}
%\begin{aligned}
%	\beta_t(i, e) &= p\left(z_{t+1}, z_{t+2}, \cdots, z_T \mid q_{[t-e+1, t]} = i, \Sigma\right) \\
%	&= p\left(z_{t+1}^{T} \mid \left(q_t, \varepsilon_t\right) = (i, e)\right).
%\end{aligned}
%\end{equation}
This formulation improves the efficiency of computing the forward and backward variables in the AHSMM, leading to reduced computational complexity compared to the traditional HSMM. In the backward variable $\beta_t(i, e)$, the quality state $i$ has been active for $e$ time steps. By summing over all quality states and potential dwell times, the backward recursion can be expressed as follows:

%This formulation allows for a more efficient computation of the forward and backward variables in the AHSMM, thereby reducing the overall computational complexity compared to the traditional HSMM.
%
%
%
%In the backward variable $\beta_t(i, e)$, the quality state $i$ has already been in the state for $e$ time steps. By aggregating across all quality states and each potential dwell time, the backward recursion can be articulated as:
%\begin{equation}
%\begin{aligned}
%	\beta_t(i, e) &= \beta_{t+1}(j, e+1) \prod_{s=1}^e b_i\left(z_{t+s}\right) + \\
%	&\quad \left(\sum_{j=1, j \neq i}^N \beta_{t+1}(j, 1) a_{i j}\right) p_i(e) b_j\left(z_{t+1}\right).  \\
%\end{aligned}
%\end{equation}

\begin{equation}   
	\begin{aligned}
	\beta_t(i, e) = \sum_{j} \left( \mathbb{I}(j \neq i) \beta_{t+1}(j, 1) a_{ij} p_i(e) b_j\left(z_{t+1}\right) \right) \\ 
	+ \beta_{t+1}(j, e+1) \prod_{s=1}^e b_i\left(z_{t+s}\right).    
    \end{aligned}
\end{equation}

We estimate the quality state $q_t$ and update the parameters of the model $\Sigma$. Using the previously defined forward and backward variables, we derive $q_t$ and adjust the model parameters. Given the model $\Sigma$ and the observation sequence $Z_{1: T}$, let $\xi_t^e(i, j)$ represent the joint probability of the observation sequence $Z_{1: T}$ and the transition from quality state $i$ to quality state $j$ (where $i \neq j$ ) at time $t$. The specific formula is as follows:

%We estimate quality state \( q_t \) and parameter re-estimation of the model \( \Sigma \). We use the forward and backward variables defined above to estimate the quality state \( q_t \) and re-estimate the parameters of the model \( \Sigma \). Given the model \( \Sigma \) and the observation sequence \( Z_{1:T} \), let \( \xi_t^e(i, j) \) denote the joint probability of the observation sequence \( Z_{1:T} \) and transitioning from quality state \( i \) to quality state \( j \) (where \( i \neq j \)) at time \( t \). The specific formula is as follows:
%\begin{equation}
%	\begin{aligned}
%		\xi_t^e(i, j) &= p\left(z_{1:T}, q_{t-1} = i, q_t = j \mid \Sigma\right) \\
%		&= \alpha_{t-1}(i, e) \left(\sum_{j=1, j \neq i}^N \beta_t(j, 1) a_{ij}\right) p_i(e) b_j\left(z_t\right).
%	\end{aligned}
%\end{equation}

\begin{equation}
\begin{aligned}
	\xi_t^e(i, j)=p\left(z_{1: T} \mid q_{t-1}=i, q_t=j, \Sigma\right) \\ \cdot p\left(q_{t-1}=i, q_t=j \mid \Sigma\right) .  
	\end{aligned}
\end{equation}

%Estimation of Health State from Observation Sequence. To estimate the quality state from the observation sequence \( Z_{1:T} \), given the model \( \Sigma \) and the observation sequence \( Z_{1:T} \), we define the joint probability \( \gamma_t^d(i) \) of being in quality state \( i \) (with an existing dwell time \( d \)) and the observation sequence \( Z_{1:T} \). Based on Equation (9), the recursion for \( \gamma_t^d(i) \) is given by:
To determine quality states from a sequence of observations, it is essential to have both a predefined model and the sequence of observations. The following equation describes the recursive estimation of quality states based on this model:

\begin{equation}
\begin{aligned}
	\gamma_t^e(i)=\sum_{j=1}^E \alpha_t(j, e) \cdot p\left(q_t=i \mid \Sigma\right) \\ \cdot p\left(z_{1: T} \mid q_t=i, \Sigma\right).
\end{aligned}
\end{equation}

%\begin{equation}
%	\begin{aligned}
%		\gamma_t^e(i) &= p\left(q_t = i \mid z_{1:T}, \Sigma\right) \\
%		&=\gamma_1^e(i)=\sum_{i=1}^E \alpha_1(i, e), t>1 .
%	\end{aligned}
%\end{equation}

This recursive formulation allows for the estimation of the quality state at time \( t \) based on the observation sequence \( Z_{1:T} \) and the given model parameters \( \Sigma \).

\section{AHSMM Prediction and Client Scheduling}

\subsection{AHSMM  Parameter Estimation}

Monitoring a device with multiple parameters can significantly improve quality prediction. Given the inherent differences among parameters, effective data fusion is essential for integrating their information. Consequently, estimating the parameters of the AHSMM becomes necessary. This estimation process utilizes Maximum Likelihood Linear Regression (MLLR) transformations to address the variations across parameters. Simultaneously, a canonical model is trained based on a set of MLLR transformations. Linear transformations are then applied to the mean vectors of the state output and dwell time distributions in the standard model, allowing for the derivation of mean vectors for these distributions.
%When a device is monitored using multiple parameters, quality prediction can achieve better performance. Due to the differences among various parameters, data fusion is required to integrate the information effectively. Therefore, it is necessary to estimate the parameters of AHSMM. The fundamental concept behind the estimation involves employing Maximum Likelihood Linear Regression (MLLR) transformations to manage variations across different parameters. At the same time, a canonical model is trained using a set of MLLR transformations. Linear transformations are applied to the mean vectors of the state output and dwell time distributions in the standard model to derive the mean vectors for these distributions.  
The formulas are given by:

\begin{equation}
	\begin{aligned}
		\begin{cases}
			b_i\left(z^{(s)}\right) = N\left(Z; \mu_i^{(s)} \Sigma_i\right),   \\
			p_i(e) = N\left(e; \mu_e^{(s)}, \sigma_i^2\right),  \quad  \\ 
			\mu_e^{(s)} = \delta^{(s)} m_i + \psi^{(s)}.        
		\end{cases}.       
	\end{aligned}
\end{equation}

Here,  $b_i\left(\boldsymbol{z}^{(s)}\right)$ represents the probability density function for the state $i$ based on the observed data $z^{(s)}$ from parameter $s$, modeled as a multivariate normal distribution with mean $\mu_i^{(s)}$ and covariance $\Sigma_i$. The term $p_i(e)$ signifies the probability density function for the dwell time $e$ in state $i$, also following a normal distribution characterized by mean $\mu_e^{(s)}$ and variance $\sigma_i^2$. The mean dwell time $\mu_e^{(s)}$ is calculated using the formula $\mu_e^{(s)}=\delta^{(s)} m_i+\psi^{(s)}$, where $\delta^{(s)}$ is a scaling factor, $m_i$ represents a parameter related to state $i$, and $\psi^{(s)}$ is a parameter-specific offset.  
%where \(\left[\boldsymbol{\eta}^{(s)}, \boldsymbol{\xi}^{(s)}\right]\) is the \( n \times (n+1) \) state output probability distribution transformation matrix for parameter \( s \), and \(\left[\boldsymbol{\delta}^{(s)}, \boldsymbol{\Psi}^{(s)}\right]\) is the \( 1 \times 2 \) state dwell probability transformation matrix for parameter \( s \). Here, \(\boldsymbol{\eta}^{(s)}\) is an \( n \times n \) matrix, \(\boldsymbol{\xi}^{(s)}\) is an \( n \)-dimensional vector, \(\left[\mu_i, 1\right]^{\mathrm{T}}\) is an \((n+1)\)-dimensional vector, and \(\left[m_i, 1\right]^{\mathrm{T}}\) is a 2-dimensional vector.

%\begin{equation}
%\begin{aligned}
%	\boldsymbol{b}_{i}\left(\boldsymbol{z}^{(s)}\right) &= \boldsymbol{N}\left(\boldsymbol{Z}; \boldsymbol{\eta}^{(s)} \boldsymbol{\mu}_i + \boldsymbol{\xi}^{(s)}, \boldsymbol{\Sigma}_i\right), \\
%	p_i(e) &= N\left(e; \delta^{(s)} m_i + \psi^{(s)}, \sigma_i^2\right),
%\end{aligned}
%\end{equation}

% Joint Estimation of AHSMM Parameters. 

% In the AHSMM, the optimal model parameter set \(\Sigma\) and the transformation matrix \(\boldsymbol{\Omega}\) are jointly estimated. Consequently, the parameter re-estimation formula for AHSMM is given by $\overline{\mu_i}, \bar{\Sigma}_i,\bar{m}_i,\bar{\sigma}_i$. 
 Let  $S$ denote the number of parameters, and let $\boldsymbol{Z} = (\boldsymbol{Z}^{(1)}, \cdots, \boldsymbol{Z}^{(S)})$ represent the monitoring data, where $\boldsymbol{Z}^{(s)} = (z_{1s}, \cdots, z_{Ts})$ represents the monitoring data of parameter $ s$ with length $T_s $. 
Here, the parameters are estimated by jointly considering the contributions of all parameters and their respective transformations. The term \(\gamma_t^e(i)\) represents the probability of being in state \( i \) with dwell time \( d \) at time \( t \), \(\eta^{(s)}\) and \(\xi^{(s)}\) are the transformation matrices and vectors for parameter \( s \), and \(\Sigma_i\) is the covariance matrix for state \( i \). This joint estimation process ensures that the model parameters are optimally adjusted for the diverse parameter data. The MAP estimation of state \( q_t \) using the AHSMM is calculated in the following way:

\begin{equation}
	\hat{q}_t=\arg \max _{\Sigma} \prod_{s=1}^S\left(p\left(Z^s \mid q_t, \Sigma\right)\right.. 
\end{equation}

%\begin{equation}
%	\hat{q}_t = \arg \max _\Sigma p(O \mid \Sigma) = \arg \max _\Sigma \prod_{s=1}^S p\left(O^s \mid \Sigma\right).
%\end{equation}

In client quality diagnostics and forecasting, parameters like computation, communication, and data quality influence decision-making differently. The  AHSMM effectively integrates these diverse parameters by capturing temporal dependencies and assigning appropriate weights. This enables accurate client quality assessment, improving forecasting and scheduling in dynamic environments.

%In client quality diagnostics and forecasting, information from various parameters often plays distinct roles in decision-making. The AHSMM addresses the challenge of effectively integrating data from multiple parameters.

\subsection{Prediction Process Based on AHSMM} 

%In practical applications, the hazard rate (HR) is often used to describe the quality status of client, utilizing the mathematical properties of the hazard rate to characterize the operational features of the client. Let \( T \) represent the lifespan of the client, \( F(t) \) denote the failure probability, and \( R(t) \) denote the reliability function, where \( F(t) + R(t) = 1 \).

 Assuming \( F(0) = 0 \) and the failure probability density function \( f(t) = F^{\prime}(t) \), the HR function is defined as:
 
 \begin{equation}
 	\Sigma(t)=\frac{f(t)}{1-F(t)}=\frac{\mathrm{d} k(t)}{N-k(t)  \mathrm{d} t} .
 \end{equation}
 A device transitions through multiple quality states before ultimately reaching a failure state. Let $E(i)$ represent the residence time in quality state $i$. We can express $E(i)$ as follows:

%\begin{equation}
%\begin{aligned}
%	\Sigma(t) &= \lim_{\substack{N \rightarrow \infty \\ \Delta \rightarrow 0}} \frac{\Delta k(t)}{[N - k(t)] \Delta t} = \\
%	&= \frac{\mathrm{d} k(t)}{[N - k(t)] \mathrm{d} t} = \frac{\mathrm{d} k(t) / M}{1 - F(t)} = \frac{f(t)}{R(t)}.
%\end{aligned}
%\end{equation}

%In this equation: \( M \) is the total number of samples, \( k(t) \) is the number of samples that have failed by time \( t \), and \( \Delta k(t) \) is the number of samples that fail in the time interval \((t, t + \Delta t)\).

$$
E(i)=m(i)+\rho \sigma^2(i),
$$
where $m(i)$ denotes the mean dwell time in state $i$ and $\sigma^2(i)$ is the variance of the dwell time in that state, and the term $\rho$ serves as a proportionality constant, which adjusts the influence of the variance on the overall residence time.

This formulation captures the idea that the total time spent in a quality state is influenced not only by the average time spent there but also by the variability of that time. A higher variance indicates greater uncertainty in the duration spent in state $i$, which can lead to longer overall residence times. By incorporating both mean and variance, we obtain a more comprehensive view of the dynamics in quality states.
%A device goes through various quality states before reaching a failure state. Let \(E(i)\) denote the residence time in the quality state \(i\). We can describe \(E(i)\) as follows:  
%
%%Before reaching its failure state, the client undergoes various quality states \( i \) \((i = 1, 2, \cdots, n-1)\). Let \( D(i) \) represent the dwell time in quality state \( i \). Based on Equations (15) and (16), \( D(i) \) can be described as:
%\begin{equation}
%E(i) = m(i) + \rho \sigma^2(i),
%\end{equation}
%where \( m(i) \) is the mean dwell time in state \( i \) and \(\sigma^2(i)\) is the variance, with \(\rho\) being a proportionality constant.
%
%This formulation allows for effective quality prediction by leveraging the characteristics of the hazard rate and the dwell time distributions in various quality states, as modeled by the AHSMM.
The proportionality constant \(\rho\) is defined as:
\begin{equation}
	\rho=\frac{T-\sum_{i=1}^N m(i)}{\sum_{i=1}^N \sigma^2(i)},   
\end{equation}
%\begin{equation}
%\rho = \left(T - \sum_{i=1}^N m(i)\right) / \sum_{i=1}^N \sigma^2(i),
%\end{equation}
where \( T \) is the total lifespan, \( m(i) \) is the mean dwell time in state \( i \), and \(\sigma^2(i)\) is the variance.

%When the client enters quality state \( i \), the Remaining Useful Life (RUL) is the sum of the remaining dwell time in the current quality state \( i \) and the dwell times in all subsequent quality states.

%Let \(\bar{D}(i, d)\) represent the remaining useful dwell time of the client in the current quality state \( i \) with an already experienced dwell time \( d \). When the client enters quality state \( i \) at time \( t \), the conditional probability of the hazard rate in the time interval \((t + d \Delta t, t + (d+1) \Delta t)\) is defined as the ratio of the probability of the client transitioning from quality state \( i \) to any other state within the future time \(\Delta t\) to the probability of the client remaining in the current quality state \( i \) at time \( t + \Delta t \).

The reliability function \( R(t + e \Delta t) \) represents the probability that the client remains in the current quality state \( i \) at time \( t + e \Delta t \). Thus, we can get
\begin{equation}
	\label{13} 
 \gamma_t^e(i)=R(t + e \Delta t).
\end{equation}
and  
\begin{equation}
	\label{14} 
	\frac{\bar{\Sigma}(t+e \Delta t)}{\gamma_t^e(i)}=\frac{\xi_t^e(i, j)}{\Delta t}.   
\end{equation}
%\begin{equation}
%\bar{\Sigma}(t + e \Delta t) \Delta t = \frac{\xi_t^e(i, j)}{\gamma_t^e(i)}.
%\end{equation}
Based on the above equations, \(\bar{E}(i, e)\) can be expressed as:
\begin{equation}
	\label{15}
	\bar{E}(i, e)=E(i)-\frac{E(i) \xi_t^e(i, j)}{\gamma_t^e(i)}. 
\end{equation}
%\begin{equation}
%\begin{aligned}
%	\bar{E}(i, e) &= E(i)(1 - \bar{\Sigma}(t + e \Delta t) \Delta t) \\
%	&= E(i)\left(1 - \frac{\xi_t^e(i, j)}{\gamma_t^e(i)}\right).
%\end{aligned}
%\end{equation}

%Using the previous equations, once the client reaches state $i$ and has a dwell time, the resulting remaining useful life (RUL) is determined as:

According to  equations (\ref{13},\ref{14},\ref{15}), once the client reaches state $i$ and has a dwell time, the trust level (TL) is determined as:
%\begin{equation}
%RUL_{(i, e)} = \bar{E}(i, e) + \sum_{j=i+1}^N E(j).
%\end{equation}

\begin{equation}
	TL_{(i, e)}=\bar{E}(i, e)+\left(\sum_{j=1}^N E(j)-\sum_{j=1}^i E(j)\right) .
\end{equation}

\subsection{Client Selection}

Considering the configuration in SD-FL, the model aggregation within the cluster at server $s$ can be described as follows:

\begin{equation}
	\boldsymbol{w}_{s,t}=\sum_{ i \in U_s}  \frac{n_i}{\sum n_i} \boldsymbol{w}_{i, t},
\end{equation}
where $U_s$ represents the set of clients assigned to server $s$, $n_i$ is the size of the local dataset for client $i$, and $\boldsymbol{w}_{i, t}$ denotes the local model of client $i$ at training round $t$. This weighted aggregation ensures that clients with larger datasets contribute proportionally more to the cluster model, improving the robustness of the overall learning process.

Furthermore, the model consensus between clusters at server $s$ during training round $t$ is characterized as follows:

\begin{equation}
	\boldsymbol{g}_{s, t}=\sum_{s=1}^S \frac{N_s}{\sum N_s} \boldsymbol{w}_{s, t},  
\end{equation}
where $N_s$ corresponds to the total size of the datasets managed by server $s$, and $S$ is the number of servers. This global aggregation step aligns the models from different clusters, ensuring consistency and convergence across the distributed system.

In the subsequent training round, each client utilizes the updated global model $\boldsymbol{g}_{s, t}$ as the initialization for their local model updates. Clients then train their local models using their respective datasets through the gradient descent mechanism, defined as follows:

\begin{equation}
	\boldsymbol{w}_{i, t+1} = \boldsymbol{w}{i, t} - \eta \nabla F_i(\boldsymbol{w}_{i, t}),
\end{equation}
where $\eta$ is the learning rate, and $\nabla F_i(\boldsymbol{w}_{i, t})$ represents the gradient of the local loss function $F_i$ with respect to the current model $\boldsymbol{w}_{i, t}$.

This decentralized training mechanism enables clients to collaboratively train a global model while keeping their data local, thereby addressing privacy concerns and minimizing the communication overhead associated with transmitting raw data. The combination of local training, cluster-level aggregation, and global consensus helps achieve a balance between computational efficiency, communication cost, and model accuracy in distributed learning systems.

\subsection{Problem Formulation}

Client scheduling determines the optimal client-server association matrix $\boldsymbol{d}$ to minimize the global model loss. Each element $d_{ij}$ is binary (1 if client $i$ is assigned to server $j$, 0 otherwise). The configuration of $\boldsymbol{d}$ directly impacts the global loss by influencing data locality, communication overhead, and computational load balancing.  Adaptive scheduling dynamically adjusts $\boldsymbol{d}$ to further enhance system performance and ensure efficient training. The global loss function is defined as:

%
%The objective of client scheduling in this context is to identify the optimal client-server association matrix $\boldsymbol{a}$, which minimizes the overall training loss function described as follows:
\begin{equation}
F(\boldsymbol{g})=\frac{1}{N} \sum_{s=1}^S \sum_{i \in \mathcal{U}_s} F_i\left(\boldsymbol{g}_s\right).
\end{equation}
%where $K$ represents the aggregate data size possessed by all clients within the system. $\boldsymbol{g}s$ denotes $\boldsymbol{g}{s, t}(\boldsymbol{a})$ for all $t$ in $T$, and $\boldsymbol{g}$ symbolizes the corresponding global model. $F_i\left(\boldsymbol{g}_s\right)$ is the local training loss for client $i$, which is defined as follows:
%%where $K$ is the total data size owned by all the clients in the system. $\boldsymbol{g}_s$ is short for $\boldsymbol{g}_{s, t}(\boldsymbol{a}), \forall t \in T . \boldsymbol{g}$ is the equivalent global model. $F_i\left(\boldsymbol{g}_s\right)$ represents the local training loss of client $i$ which is given by
%\begin{equation}
%F_i\left(\boldsymbol{g}_s\right)=\sum_{k=1}^{K_i} f\left(\boldsymbol{g}_s, \boldsymbol{x}_{i k}, \boldsymbol{y}_{i k}\right), i \in \mathcal{U}_s .
%\end{equation}

Therefore, the optimization of the client-server association matrix can be achieved by solving the problem of minimizing the global training loss:

%The global training loss function is typically defined as the weighted sum of the training losses of all clients, where the weights may be related to the volume or quality of each client's data. In this optimization problem, constraints must be considered to ensure that each client is associated with only one server, the processing capacity of the servers is not exceeded, and bandwidth and latency limits of the network may also need to be considered. To solve this problem, linear programming, integer programming, or greedy algorithms can be used. However, they need to be adjusted based on the actual network environment and server capabilities to ensure the efficiency of data transmission and processing.

%\begin{equation}
%\begin{aligned}
%	\min_{\boldsymbol{d}} & F(\boldsymbol{g})=\frac{1}{N} \sum_{s=1}^S \sum_{i \in \mathcal{U}_s} F_i\left(\boldsymbol{g}_s\right) \\
%	\text{subject to} & \quad \sum_{\forall s \in \mathcal{S}} d_{i, s} \leq 1,   \\
%	&  \quad d_{i, s} \in \{0,1\}, \\
%	& \quad TL_i \geq \Theta.  
%\end{aligned}
%\end{equation}
%
%
%
%
%\begin{equation}
%	\begin{aligned}
%		\min_{\boldsymbol{d}} & \quad F(\boldsymbol{g}) = \frac{1}{N} \sum_{s=1}^S \sum_{i \in \mathcal{U}_s} F_i\left(\boldsymbol{g}_s\right) \\
%		\text{subject to} & \quad \sum_{\forall s \in \mathcal{S}} d_{i, s} \leq 1,   \\
%		& \quad d_{i, s} \in \{0,1\}, \\
%		& \quad TL_i \geq \Theta.  
%	\end{aligned}
%\end{equation}

\begin{equation}
	\begin{aligned}
		\min_{\boldsymbol{d}} & \quad F(\boldsymbol{g}) = \frac{1}{N} \sum_{s=1}^S \sum_{i \in \mathcal{U}_s} F_i\left(\boldsymbol{g}_s\right) \\
		\text{subject to} & \quad \sum_{\forall s \in \mathcal{S}} d_{i, s} \leq 1,   \\
		& \quad d_{i, s} \in \{0,1\}, \\
		& \quad TL_{i,s} \geq \Theta.  
	\end{aligned}
\end{equation}

%where restricts the client-server association indicator, (10b) implies that each client can associate with no more than one edge server at the same time, and (10c) restricts the maximum number of clients (i.e., $U_{\max }$ ) that one edge server can associate with. Problem (10) is hard to solve due to the implicit relationship between the objective function and the optimization variable. In the subsequent section, we thus conduct the convergence analysis of the FL system to make such a relationship explicit and use the analytical results for high-efficient client scheduling algorithm design.

\subsection{Convergence Analysis}

We rely on current trust levels of client $i$  to tackle these challenges, reduce data loss, and guarantee consistent model updates. $\Theta$ 
denotes the threshold for the trust level of clients involved in training. Reformulating the optimization problem to incorporate parameters that reflect communication link stability will enhance the modeling of the SD-FL system's conditions.

We also plan to enhance SD-FL system robustness through redundancy strategies, like multiple communication paths or backup servers, to mitigate risks from unreliable links. Dynamically adjusting client-server associations based on real-time assessments will help maintain optimal performance despite trust fluctuations. This approach maximizes resource utilization and minimizes training time, leading to more robust convergence and broader adoption in real-world applications.

\begin{theorem}      	
	By setting the learning rate $\lambda = \frac{1}{L}$, the upper bound of the expected difference $\mathbb{E}\left(F\left(\boldsymbol{g}_{t+1}\right) - F\left(\boldsymbol{g}^*\right)\right)$ can be established as follows:
%	Setting the learning rate to $\lambda=\frac{1}{L}$ allows us to establish the upper limit of $\mathbb{E}\left(F\left(\boldsymbol{g}_{t+1}\right)-F\left(\boldsymbol{g}^*\right)\right)$ in the following manner.  
%	The expected difference between the function value at the updated iterate $\boldsymbol{g}_{t+1}$ and the optimal function value $\boldsymbol{g}^*$ can be bounded as shown below:
	\begin{equation}
		\begin{aligned}
			\mathbb{E}\left(F\left(\boldsymbol{g}_{t+1}\right)-F\left(\boldsymbol{g}^*\right)\right) \leq & D^t \mathbb{E}\left(F\left(\boldsymbol{g}_0\right)-F\left(\boldsymbol{g}^*\right)\right) \\
			& +\frac{2 \omega_1 B}{L} \frac{1-D^t}{1-D}
		\end{aligned},
	\end{equation}
	%\begin{equation}
	%\mathbb{E}\left(F\left(\boldsymbol{g}_{t+1}\right)-F\left(\boldsymbol{g}^*\right)\right) \leq A^t \mathbb{E}\left(F\left(\boldsymbol{g}_0\right)-F\left(\boldsymbol{g}^*\right)\right)+\frac{2 \omega_1 B}{L} \frac{1-A^t}{1-A}
	%\end{equation}
	where 
$
		D=1-\frac{\mu}{L}+\frac{4 \omega_2 \mu B}{L}
	$
	, $	B=\sum_{m \in \mathcal{S}} \frac{1}{N_m^{(S)}}\Psi$ ,  and $  \Psi=\left(\sum_{i \in \mathcal{U}} n_i d_{i, m}\left(D_m-1+\mathbb{I}\left(TL_{i,s}< \Theta \right)\right) \right)$.   
%	\begin{equation}
%		\begin{cases}
%%			\sum_{s \in \mathcal{S}} \sum_{m \in \mathcal{S}} \frac{1}{K_m^{(S)}}\left(\sum _ { i \in \mathcal { U } } K _ { i } a _ { i , m } \left(\frac{1-p_{s, m}}{1+\sum_{j \in \mathcal{S}_s}\left(1-p_{s, j}\right)}\right.\right. \\
%%			\left.\left.-\frac{1}{S}\left(1-q_i\left(a_{i, m}\right)\right)\right)\right) \\
%			B=\sum_{m \in \mathcal{S}} \frac{1}{N_m^{(S)}}\Psi,    \\
%			\Psi=\left(\sum_{i \in \mathcal{U}} n_i d_{i, m}\left(D_m-1+\mathbb{I}\left(TL_{i,s}< \Theta \right)\right) \right).
%		\end{cases}
%	\end{equation}
	
\end{theorem}

In the definition of $B$, the following equations hold:
\begin{equation}
	N_m^{(S)}=\sum_{i \in \mathcal{U}} n_i d_{i, m},
\end{equation}
and
\begin{equation}
	D_m= \frac{1}{1+|S|} .
\end{equation}

\begin{algorithm}
	\caption{Greedy Algorithm for Solving the Optimization Problem}
	\label{alg:greedy_optimization}
	\KwIn{Set of clients $\mathcal{U}$, set of servers $\mathcal{S}$, maximum clients per server $U_{\max}$, the allocation status $x_i$; }
	\KwOut{Assignment matrix $\boldsymbol{d}$;}
	Initialize $\boldsymbol{d}_{i,s} = 0, \forall i \in \mathcal{U}, \forall s \in \mathcal{S}$\;
	Initialize client count $u_s = 0, \forall s \in \mathcal{S}$\;
	Sort  $(i,s) \in (\mathcal{U},\mathcal{S})$ in increasing order based on $TL_{i,s}$  for each server $s$\;
	\For{(i,s) in $(\mathcal{U},\mathcal{S})$}{
			\If{$u_s < U_{\max}$  and $x_i=0$ and $TL_{i,s}\geq \Theta$ }{
				Assign client $i$ to server $s$: $d_{i, s} = 1$\;   
				Increment client count for server $s$: $u_s = u_s + 1$\; 
				$x_i=1$;      
			}
		
	}
%	\Return $\boldsymbol{d}$\;
\end{algorithm}

%We note that is capped at $\frac{2 \omega_1 B}{L(1-A)}$ as $t$ mechanismes infinity and $A$ is negative. This upper bound demonstrates the gap between the converging and optimal global training losses, prompting us to reduce it to improve loss minimization by targeting $\ \ \frac {2 \omega_1 B}{1-A}$ and ultimately $B$. $B$ is related to the Packet Error Rate (PER) in communications, linking optimization variable $a$ to global training loss and underscoring the impact of unreliable communications on learning outcomes.

From Theorem 1, the global training loss minimization initially outlined in the problem can be reinterpreted to focus primarily on minimizing the parameter $B$. This revised formulation, therefore, positions $B$ as the central target for reduction, aiming to directly influence and improve the overall system performance by addressing the underlying factors contributing to $B$ 's value, i.e., 
\begin{equation}
	\begin{aligned}
		\min _d & \sum_{m \in \mathcal{S}} \frac{1}{N_m^{(S)}}\left(\sum_{i \in \mathcal{U}} n_i d_{i, m}\left(D_m-1+\mathbb{I}\left(TL_i< \Theta \right)\right) \right). \\
	\end{aligned}
\end{equation}
To address this nonlinear integer programming problem, we propose a greedy algorithm outlined in Algorithm 1, with a time complexity of $O(nm\log nm)$.

\section{ Experiments Evaluation}

%This section provides a evaluation of our proposed mechanism using four real-world datasets to demonstrate its effectiveness and practicality. We begin by introducing the experimental setting. Following this, we present the results of our experimental comparisons, highlighting the performance of our mechanism relative to existing state-of-the-art approaches.

This section evaluates our proposed mechanism using  real-world datasets to demonstrate its effectiveness and practicality. We begin by introducing the experiment setting. Then, we present our experimental comparisons' results, highlighting our mechanism's performance relative to the baselines.

%This section will evaluate our proposed mechanism using four real-world datasets. First, we will introduce the experiment setting. Then, we will present the results of our experimental comparisons.

\subsection{Experments Setting}

 We provide a detailed explanation of the fundamental experimental setup, including the basic setup,  datasets, training configurations,   baselines, and evaluation metrics.

\textbf{Basic Setup: }
We design a SD-FL system comprising five edge servers and fifty clients, with each client assigned 1,000 local training samples.  To emulate real-world challenges, 10\%, 30\%, and 50\% of the clients gradually experience degradation in both training quality (e.g., training  accuracy) and communication quality (e.g.,  packet loss) as training progresses.

%	We set up a system with five edge servers and fifty clients. Each client possesses 1,000 local training data samples. Additionally, we have set up the system so that 10\%, 30\%, and 50\% of the clients experience a gradual decline in training quality and communication quality as the training progresses. 

\textbf{Datasets:} Real-world Datasets.   Four  standard real-world datasets, e.g. MNIST \cite{mnist}, EMNIST \cite{emnist}, SVHN \cite{svhn},and   CIFAR-10 \cite{cifar} are utilized to make performance evaluation.

\textbf{Training Configurations:}  Training Parameters. We adopt a CNN architecture for its effectiveness in image processing tasks. The batch size is 32, balancing computational efficiency and model performance. Each client performs 100 local training rounds ($T_1 = 100$) before aggregation, with 100 inter-cluster aggregations ($T_2 = 100$) to synchronize updates across edge servers. The learning rate ($\eta$) is set to 0.01, ensuring stable and efficient optimization, and SGD with a momentum of 0.05. The model uses ReLU as the activation function for non-linearity and cross-entropy loss for classification tasks.

%	Training  Parameters.  We utilize a CNN training architecture. The batch size is 32, The system is configured for 100 local training rounds ($T_1=100$) per client, complemented by 100 inter-cluster aggregations ($T_2 = 100$) to synchronize updates across edge servers efficiently. The learning rate $\eta$ is configured to 0.01 and the SGD momentum is set to 0.05.

\textbf{Baselines:} In order to validate the effectiveness of our proposed mechanism, we compared our mechanism with the following three mechanisms.

\begin{itemize}
	\item  GUROBI:   In \cite{gurobi}, the authors utilize the GURUBI optimizer for the client's optimal allocation problem. The prediction part of the front end does not utilize the prediction mechanism of AHSMM. 
	\item  TRUST. In \cite{Wang2024TrustAoIAwareCO}, the authors introduce a trust-age of information (AoI)-aware co-design scheme (TACS), employing a learning-based trust model and trust-AoI-aware scheduling to optimize data selection for plant control dynamically.  
	
	\item  RANDOM. Here, we continue to use the AHSMM for the prediction component, while employing random allocation for client assignments.
\end{itemize}

\textbf{Evaluation metrics:} We use two metrics to evaluate our mechanisms: test accuracy and training loss. The results are obtained from the average of multiple experiments.

\begin{itemize}

	\item  Test accuracy. Test accuracy measures a model's performance on unseen data, reflecting its generalization ability and effectiveness in SD-FL, critical for real-world usability and reliability.

%	 Test accuracy is a crucial metric for model training in FL, representing the performance that model owners can achieve with their trained models. 

		\item  Training loss.  Training loss quantifies the discrepancy between the predicted outputs of a model and the actual data, guiding the optimization process to improve model accuracy and performance.
\end{itemize}

\begin{figure*}[!t]   % * 表示忽略单行
	\centering            
	
	\begin{minipage}{0.24\textwidth}
		
		\includegraphics[width=1\textwidth]{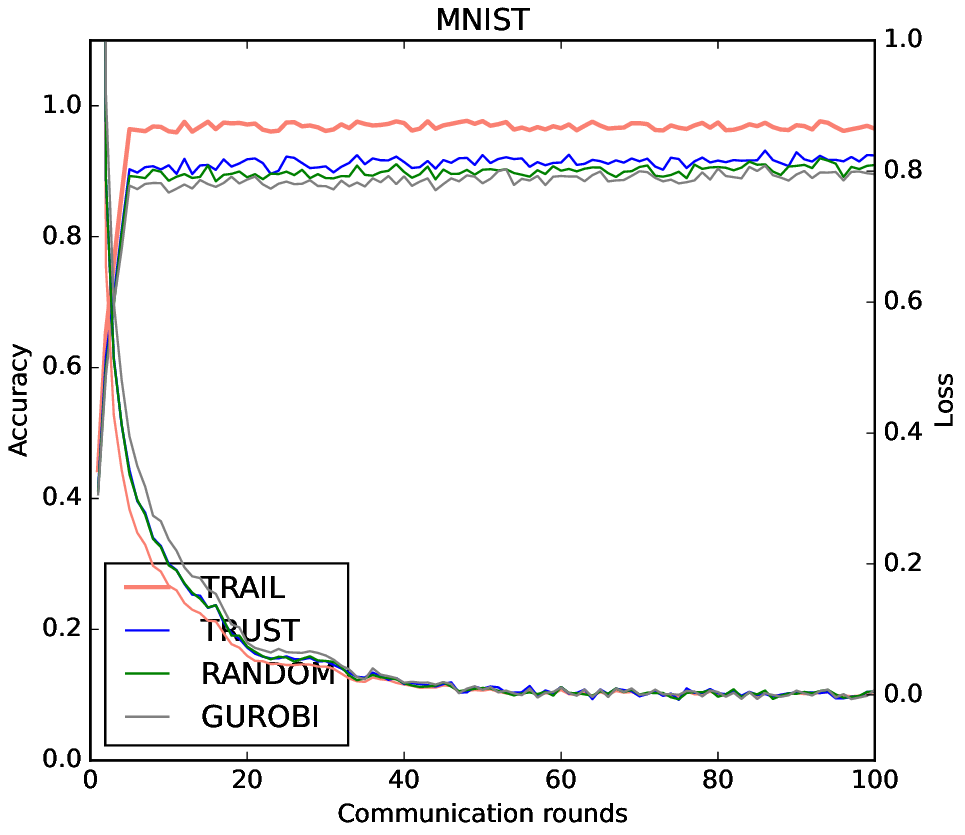}
		\centerline{(\emph{a})}
		
	\end{minipage}
	%  \hfill
	\begin{minipage}{0.24\textwidth}
		\includegraphics[width=1\textwidth]{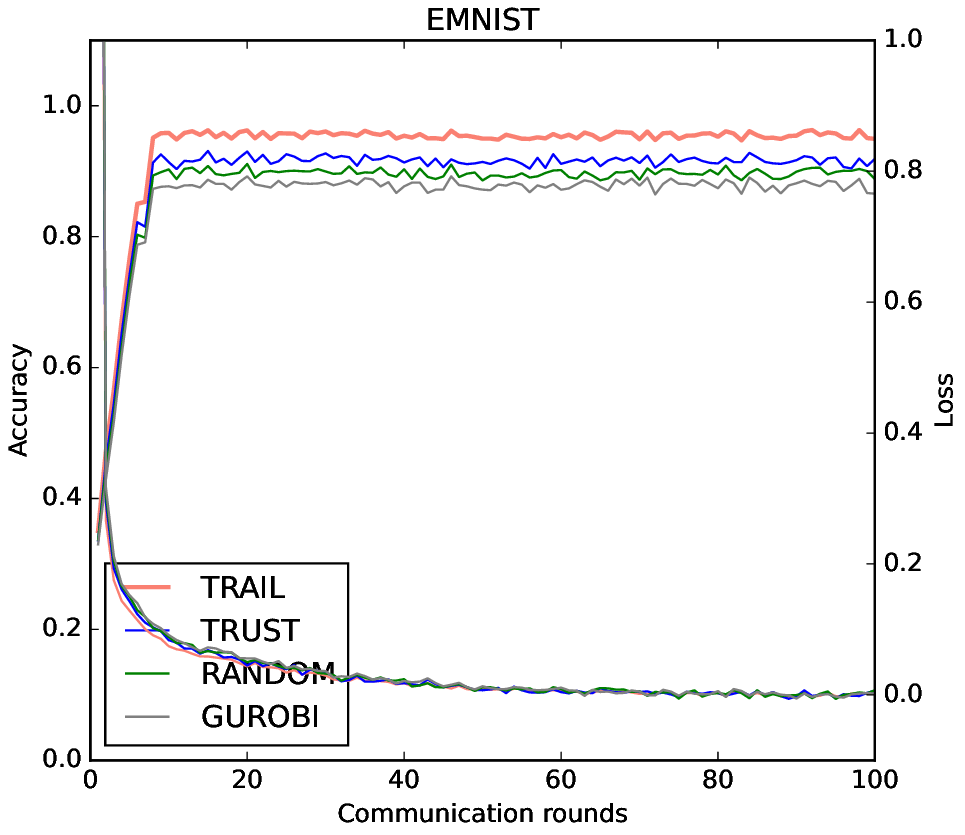}
		\centerline{(\emph{b})}
		
	\end{minipage}
	\begin{minipage}{0.24\textwidth}
		\includegraphics[width=1\textwidth]{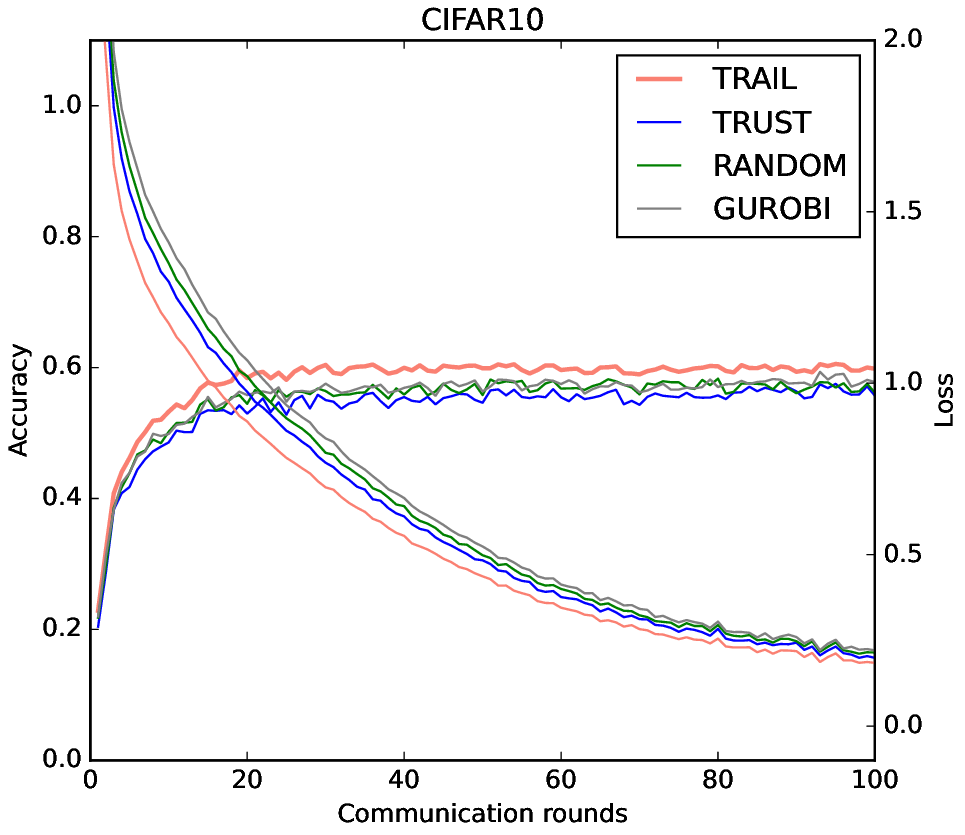}
		\centerline{(\emph{c})}
		
	\end{minipage}
	\begin{minipage}{0.24\textwidth}
		\includegraphics[width=1\textwidth]{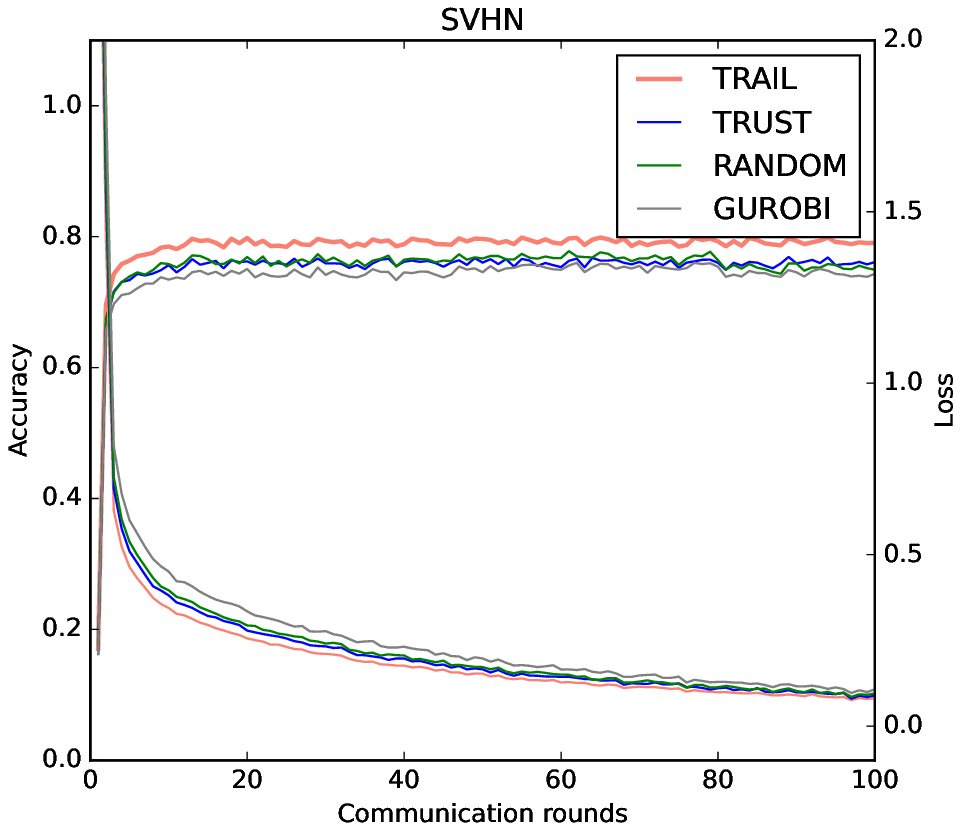}
		\centerline{(\emph{d})}
		
	\end{minipage}
	\caption{The test accuracy and training loss in scenarios with 10\% low-quality clients:  (a) MNIST, (b) EMNIST, (c) CIFAR10, and (d) SVHN.  } 
	\label{simulateddata}       
\end{figure*}

\begin{figure*}[!t]   % * 表示忽略单行
	\centering            
	
	\begin{minipage}{0.24\textwidth}
		
		\includegraphics[width=1\textwidth]{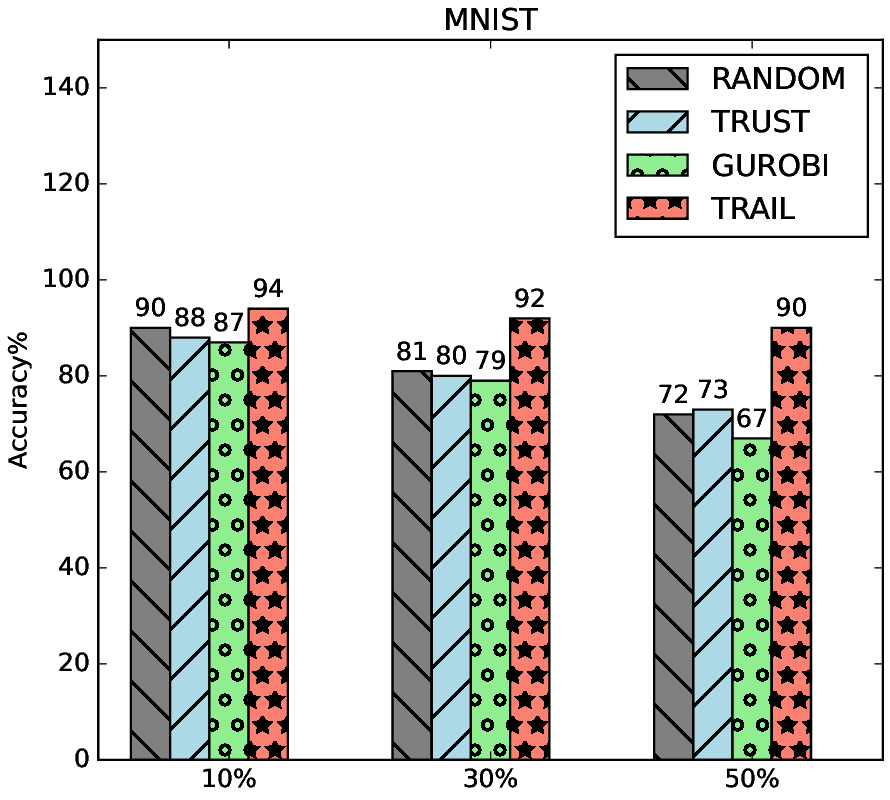}
		\centerline{(\emph{a})}
		
	\end{minipage}
	%  \hfill
	\begin{minipage}{0.24\textwidth}
		\includegraphics[width=1\textwidth]{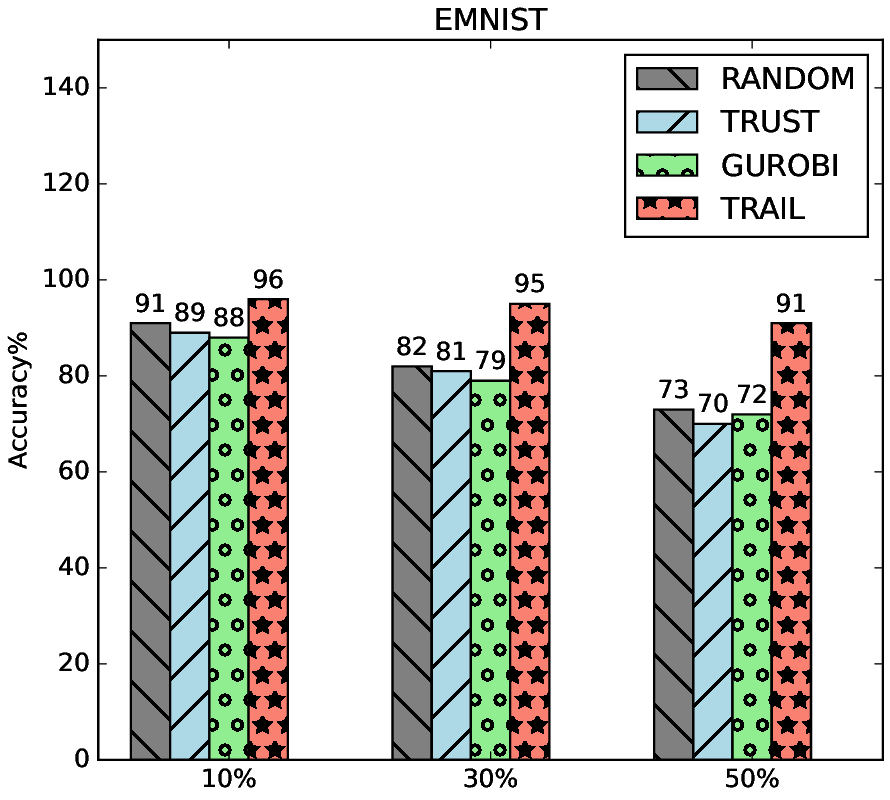}
		\centerline{(\emph{b})}
		
	\end{minipage}
	\begin{minipage}{0.24\textwidth}
		\includegraphics[width=1\textwidth]{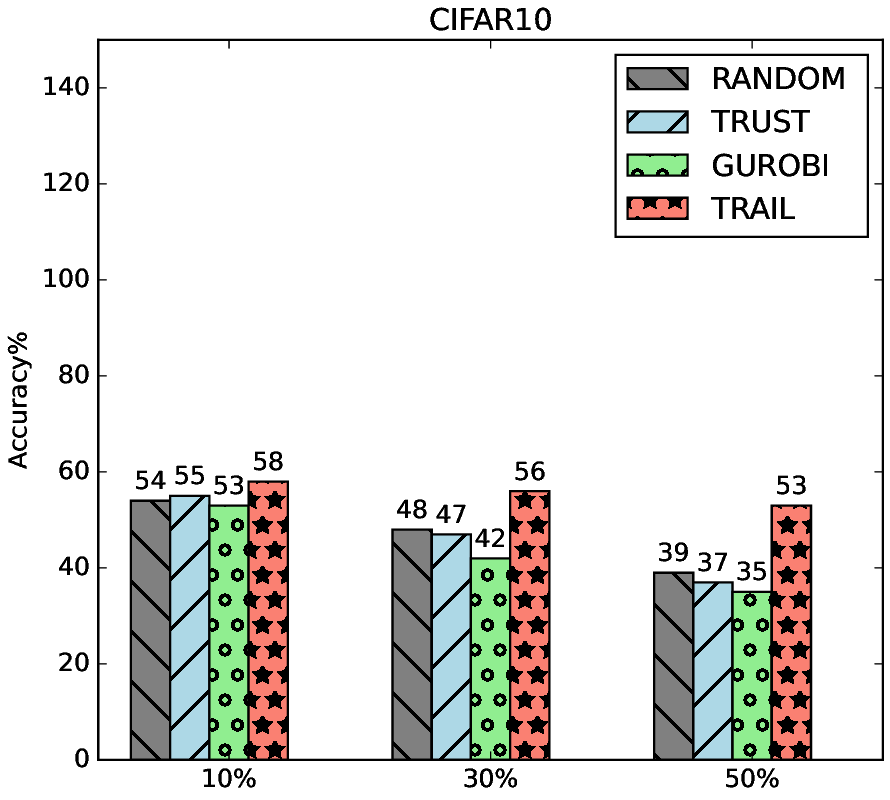}
		\centerline{(\emph{c})}
		
	\end{minipage}
	\begin{minipage}{0.24\textwidth}
		\includegraphics[width=1\textwidth]{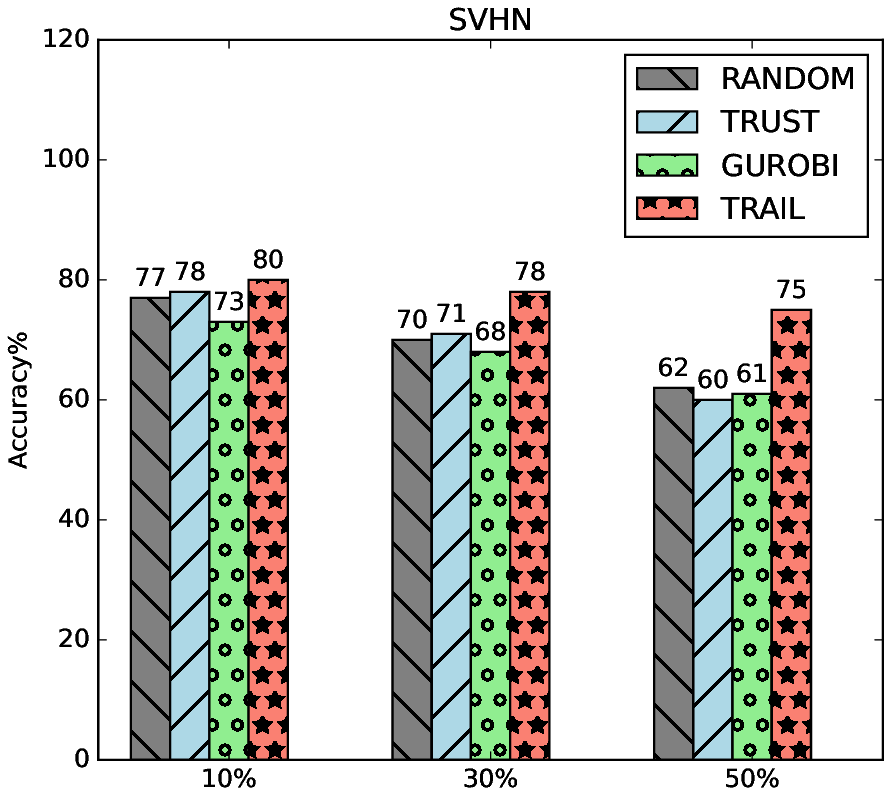}
		\centerline{(\emph{d})}
		
	\end{minipage}
	\caption{The test accuracy in scenarios with 10\%,30\%,50\% low-quality clients:  (a) MNIST, (b) EMNIST, (c) CIFAR10, and (d) SVHN.  } 
	\label{fig222}       
\end{figure*}

\subsection{Experiments Results}

%\begin{table*}[ht]
%	\centering
%	\caption{Social utility table with varying EDs and MOs}
%	\footnotesize
%	%\begin{tabular}{cccccccccccc}
%	\begin{tabularx}{\linewidth}{c*{10}{>{\centering\arraybackslash}X}}
%		\toprule
%		\multirow{2}{*}{\textbf{EDs}} & \multicolumn{10}{c}{\textbf{MOs}} \\ % 使用 multirow 调整 Clients 的位置
%		\cmidrule(lr){2-11}
%		& \textbf{2} & \textbf{4} & \textbf{6} & \textbf{8} & \textbf{10} & \textbf{12} & \textbf{14} & \textbf{16} & \textbf{18} & \textbf{20} \\
%		\midrule
%		\textbf{5}  & 8.31 & 14.21 & 18.81 & 22.10 & 25.15 & 27.44 & 29.23 & 30.86 & 31.85 & 33.32 \\
%		\textbf{10} & 16.68 & 25.09 & 29.82 & 33.44 & 35.89 & 37.84 & 39.29 & 39.62 & 40.69 & 41.54 \\
%		\textbf{15} & 23.00 & 33.34 & 39.44 & 43.19 & 45.30 & 46.66 & 47.09 & 48.26 & 48.07 & 49.25 \\
%		\textbf{20} & 28.66 & 39.60 & 46.34 & 50.12 & 52.62 & 54.36 & 56.01 & 56.65 & 57.39 & 57.82 \\
%		\textbf{25} & 33.14 & 45.83 & 51.46 & 55.23 & 58.31 & 61.24 & 62.58 & 63.32 & 64.26 & 64.86 \\
%		\textbf{30} & 37.53 & 49.94 & 56.56 & 60.39 & 63.75 & 64.85 & 67.21 & 68.08 & 68.56 & 70.26 \\
%		\textbf{35} & 41.07 & 53.88 & 60.69 & 63.46 & 67.06 & 68.60 & 69.81 & 71.07 & 72.93 & 74.18 \\
%		\textbf{40} & 44.27 & 57.35 & 64.10 & 66.19 & 69.54 & 71.39 & 73.92 & 73.69 & 74.87 & 76.92 \\
%		\textbf{45} & 47.69 & 60.10 & 67.09 & 69.79 & 72.77 & 73.99 & 75.10 & 77.10 & 77.83 & 78.19 \\
%		\textbf{50} & 49.94 & 62.21 & 69.34 & 71.86 & 75.34 & 77.42 & 77.77 & 79.35 & 79.96 & 80.94 \\
%		\bottomrule
%	\end{tabularx}
%	\label{tab2}
%\end{table*}

In Figure \ref{simulateddata}, we analyze the variations in test accuracy and training loss over multiple training rounds for four distinct mechanisms, specifically under conditions where only 10\% of clients are classified as low quality. Our proposed mechanism stands out by achieving the highest performance across four real datasets. This result is due to the effective integration of our AHSMM model with a greedy algorithm, which works together to predict fluctuations in client learning and communication quality reasonably. 
By optimizing the participation of low-quality clients, our mechanism significantly enhances overall training outcomes.
In contrast, the TRUST mechanism, despite employing a predictive approach, lacks an effective client distribution strategy. This deficiency leads to suboptimal performance, as it fails to adaptively manage client participation based on their quality. 
Similarly, the RANDOM mechanism incorporates the AHSMM to forecast client behavior, but it does not allocate clients efficiently, leading to less effective training sessions.
Although the GUROBI mechanism can determine the optimal client allocation scheme, it does not incorporate client quality predictions, which hampers its ability to promptly exclude low-quality clients from participating in training, ultimately affecting training efficiency.
Overall, our mechanism demonstrates a superior ability to navigate the complexities of client quality by leveraging predictive modeling and strategic allocation, ensuring robust training performance in SD-FL environments.

%In Figure  \ref{simulateddata}, we examine the changes in test accuracy and training loss over training rounds for four different mechanisms under the condition that only 10\% of clients are of low quality. Our proposed mechanism achieved the highest performance across four real datasets. This success is primarily due to our designed AHSMM mechanism and greedy algorithm, which effectively predict changes in client learning and communication quality, optimizing low-quality clients' participation and significantly enhancing training outcomes. In contrast, although the TRUST mechanism employs a predictive mechanism, it lacks an effective client distribution strategy, resulting in poorer performance. Similarly, while the RANDOM mechanism uses the AHSMM mechanism to predict client behavior, it fails to allocate clients efficiently. The GUROBI mechanism can determine the optimal client distribution scheme, but the solution process is time-consuming and not as efficient as our proposed greedy algorithm. Moreover, GUROBI does not predict client quality, failing to exclude low-quality clients from training participation timely.

In Figure \ref{fig222}, we analyze the comparative performance of four mechanisms across scenarios characterized by varying proportions of low-quality clients. As the percentage of low-quality clients increases, all mechanisms demonstrate a decline in both training and testing accuracy, albeit to different extents. Notably, our proposed mechanism consistently delivers the best results across four real datasets, showcasing its robustness in challenging environments.
The superior performance of our mechanism can be attributed to the integration of the AHSMM and a greedy-based client allocation algorithm. This combination effectively predicts fluctuations in client learning and communication quality, enabling efficient client allocation that significantly enhances model training quality, even under adverse conditions. By optimizing the participation of higher-quality clients, we mitigate the negative impact of low-quality clients on overall performance.
In contrast, while the TRUST mechanism is capable of identifying unreliable clients, it lacks an efficient client distribution strategy. This limitation results in poorer training outcomes compared to our mechanism, as it fails to adaptively manage client participation based on their quality. Similarly, the RANDOM mechanism employs the AHSMM to accurately predict changes in client training quality but relies on a random allocation strategy during the client distribution phase. Consequently, this randomness undermines the final training effectiveness, leading to suboptimal results.
The GUROBI mechanism, despite its potential for determining optimal client distributions, performs the worst in our experiments. Its inability to accurately predict changes in client learning quality restricts its effectiveness, as it cannot exclude low-quality clients in a timely manner. Experiments reveal a notable 8.7\% increase in test accuracy and a 15.3\% reduction in training loss compared to existing baselines, demonstrating the superiority of our mechanism in SD-FL settings. These results underscore the importance of both predictive modeling and strategic client allocation in achieving high-quality training outcomes.

%In Figure \ref{fig222}, we examine the comparative performance of four mechanisms in scenarios with varying proportions of low-quality clients. As the proportion of low-quality clients increases, all mechanisms exhibit varying degrees of decline in training and testing accuracy. However, our proposed mechanism consistently achieves the best results across four real datasets. This superior performance is attributed to our AHSMM mechanism and the greedy-based client allocation scheme, which effectively predicts changes in client learning and communication quality and efficiently allocates clients, significantly enhancing model training quality under adverse conditions. In contrast, while the TRUST mechanism can also predict unreliable clients, it lacks an efficient client distribution strategy, leading to poorer training outcomes than our mechanism. Although utilizing the AHSMM mechanism to predict changes in client training quality accurately, the RANDOM mechanism employs a random allocation strategy during the client distribution phase, thus diminishing the final training effectiveness. GUROBI performs the worst, as it fails to predict changes in client learning quality accurately. Our experiments show an 8.7\% increase in test accuracy and a 15.3\% reduction in training loss compared to existing baselines.

\section{Conclusion}  
%We designed the TRAIL (Trust-Aware Client Scheduling) mechanism within a semi-decentralized SD-FL framework, incorporating an adaptive hidden semi-Markov model (AHSMM) to accurately estimate clients' states and contributions alongside a client-server association optimization problem to minimize global training loss. As a result, this mechanism significantly enhanced model training efficiency and robustness.   

This paper proposes TRAIL, a novel mechanism designed to address the dynamic challenges in SD-FL. TRAIL integrates an AHSMM to accurately predict client states and contributions and a greedy algorithm to optimize client-server associations, effectively minimizing global training loss. Through convergence analysis, the impact of client-server relationships on model convergence is theoretically assessed. Extensive experiments conducted on four real-world datasets demonstrate that TRAIL improves test accuracy and training loss, significantly outperforming state-of-the-art baselines. This work highlights the potential of combining predictive modeling and strategic client allocation to enhance efficiency, robustness, and performance in distributed learning systems.

%In this paper, we proposed TRAIL, a novel mechanism designed to address the dynamic challenges in  SD-FL. TRAIL significantly improves training efficiency and robustness by integrating an AHSMM to predict client states and contributions and a greedy algorithm to optimize client-server associations. Experimental results on real-world datasets show an improvement compared to state-of-the-art baselines. 
%

\section{Acknowledgments}   
This work was supported in part by the National Natural Science Foundation of China under Grants 62372343, 62072411, and 62402352, in part by the Zhejiang Provincial Natural Science Foundation of China under Grant LR21F020001, in part by the Key Research and Development Program of Hubei Province under Grant 2023BEB024, and in part by the Open Fund of Key Laboratory of Social Computing and Cognitive Intelligence (Dalian University of Technology), Ministry of Education under Grant SCCI2024TB02.

\bibliography{aaai25}

\section{Appendix}

\subsection{Proof of Theorem 1}

Assumption 1: Strong Convexity and Lipschitz Continuity
Assume the global loss function $ \mathrm{F}(g) $ is $\mu$-strongly convex and its gradient is $L$-Lipschitz continuous. This implies that for any $g$ and $g'$:

\begin{itemize}
	\item Strong convexity:
	
	$$
	F\left(g^{\prime}\right) \geq F(g)+\left\langle\nabla F(g), g^{\prime}-g\right\rangle+\frac{\mu}{2}\left\|g^{\prime}-g\right\|^2.
	$$
	
    \item Gradient Lipschitz continuity:
    
    $$
    \left\|\nabla F(g)-\nabla F\left(g^{\prime}\right)\right\| \leq L\left\|g-g^{\prime}\right\|.
    $$
\end{itemize}

Assumption 2: Assume that during training, randomness is introduced (e.g., due to random client participation), and these random factors are independent and identically distributed (i.i.d.).

Definitions: Let $g_t$ denote the global model at iteration $t$ , and $g^*$ denote the global optimal model. Let $omega_1$ and $\omega_2$ be constants related to the system, and $ B$ is a cumulative error term. In FL, the global model update can be expressed as:
$$
g_{t+1}=g_t-\lambda \nabla F\left(g_t\right)+\eta_t,
$$
where $\eta_t$ represents the noise or error term due to factors like client sampling and unreliable communication.
Since $\lambda = \frac{1}{L}$, we have:
$$
g_{t+1}=g_t-\frac{1}{L} \nabla F\left(g_t\right)+\eta_t.
$$
Consider the difference:
$$
F\left(g_{t+1}\right)-F\left(g^*\right) .
$$

Using the $L$-Lipschitz continuity of the gradient, we have:
$$
F\left(g_{t+1}\right) \leq F\left(g_t\right)+\left\langle\nabla F\left(g_t\right), g_{t+1}-g_t\right\rangle+\frac{L}{2}\left\|g_{t+1}-g_t\right\|^2
$$
Substituting $9_\{t+1\} - g_t$, we get:
$$
g_{t+1}-g_t=-\frac{1}{L} \nabla F\left(g_t\right)+\eta_t,
$$
so
$$
\left\|g_{t+1}-g_t\right\| \leq \frac{1}{L}\left\|\nabla F\left(g_t\right)\right\|+\left\|\eta_t\right\| .
$$
Substitute back into the inequality:
$$
\begin{aligned}
F\left(g_{t+1}\right) \leq F\left(g_t\right)-\frac{1}{L}\left\|\nabla F\left(g_t\right)\right\|^2+\left\langle\nabla F\left(g_t\right),  \eta_t\right\rangle   \\ +\frac{L}{2}\left(\frac{1}{L}\left\|\nabla F\left(g_t\right)\right\|+\left\|\eta_t\right\|\right)^2 .
\end{aligned}
$$ 
Take the expectation of both sides. Assuming that $\eta_t $ has zero mean (the noise is mean zero), we have:

$$
\begin{aligned}
E\left[F\left(g_{t+1}\right)\right] \leq E\left[F\left(g_t\right)\right]-\frac{1}{L} E\left[\left\|\nabla F\left(g_t\right)\right\|^2\right]+ \\ \frac{L}{2} E\left[\left(\frac{1}{L}\left\|\nabla F\left(g_t\right)\right\|+\left\|\eta_t\right\|\right)^2\right] .
\end{aligned}
$$

Expanding the squared term:

$$
\begin{aligned}
\left(\frac{1}{L}\left\|\nabla F\left(g_t\right)\right\|+\left\|\eta_t\right\|\right)^2=\left(\frac{1}{L}\left\|\nabla F\left(g_t\right)\right\|\right)^2+ \\ 2 \frac{1}{L}\left\|\nabla F\left(g_t\right)\right\|\left\|\eta_t\right\|+\left\|\eta_t\right\|^2.  
\end{aligned}
$$

Therefore, we can get 
%$$
%\begin{aligned}
%E\left[F\left(g_{t+1}\right)\right] \leq E\left[F\left(g_t\right)\right]-\frac{1}{L} E\left[\left\|\nabla F\left(g_t\right)\right\|^2\right]+ \\
%\frac{L}{2}\left(\left(\frac{1}{L}\right)^2 E\left[\left\|\nabla F\left(g_t\right)\right\|^2\right]+2 \frac{1}{L} E\left[\left\|\nabla F\left(g_t\right)\right\|\left\|\eta_t\right\|\right]+E\left[\left\|\eta_t\right\|^2\right]\right) .
%\end{aligned}
%$$
%
%
%
%Simplifying:
$$
\begin{aligned}
E\left[F\left(g_{t+1}\right)\right] \leq E\left[F\left(g_t\right)\right]- \\ \left(\frac{1}{L}-\frac{1}{2 L}\right) E\left[\left\|\nabla F\left(g_t\right)\right\|^2\right]+ \\  
L E\left[\left\|\nabla F\left(g_t\right)\right\|\left\|\eta_t\right\|\right]+\frac{L}{2} E\left[\left\|\eta_t\right\|^2\right] .
\end{aligned}
$$
Note that $\frac{1}{L} - \frac{1}{2L} = \frac{1}{2L}$.
Thus,
$$
\begin{aligned}
E\left[F\left(g_{t+1}\right)\right] \leq E\left[F\left(g_t\right)\right]-\frac{1}{2 L} E\left[\left\|\nabla F\left(g_t\right)\right\|^2\right]+ \\ L E\left[\left\|\nabla F\left(g_t\right)\right\|\left\|\eta_t\right\|\right]+\frac{L}{2} E\left[\left\|\eta_t\right\|^2\right] .
\end{aligned}
$$
Since $F(g)$ is $\mu$-strongly convex, we have:
$$
\begin{aligned}
	\left\|\nabla F\left(g_t\right)\right\|^2 \geq 2 \mu\left(F\left(g_t\right)-F\left(g^*\right)\right)\\
	E\left[F\left(g_{t+1}\right)\right] \leq E\left[F\left(g_t\right)\right]-\frac{\mu}{L} E\left[F\left(g_t\right)-F\left(g^*\right)\right]+ \\ L E\left[\left\|\nabla F\left(g_t\right)\right\|\left\|\eta_t\right\|\right]+\frac{L}{2} E\left[\left\|\eta_t\right\|^2\right.
\end{aligned}
$$

Introduce constants $\omega_1$ and $\omega_2$ satisfying:
$$
E\left[\left\|\nabla F\left(g_t\right)\right\|\left\|\eta_t\right\|\right] \leq \omega_1 E\left[F\left(g_t\right)-F\left(g^*\right)\right]+\omega_2 B,
$$
and
$$
E\left[\left\|\eta_t\right\|^2\right] \leq B
$$
where $ B $ is the cumulative error term defined earlier.
Substitute these into the inequality:
$$
\begin{aligned}
E\left[F\left(g_{t+1}\right)\right] \leq E\left[F\left(g_t\right)\right]- \\ \left(\frac{\mu}{L}-L \omega_1\right) E\left[F\left(g_t\right)-F\left(g^*\right)\right]+\left(L \omega_2+\frac{L}{2}\right) B .
\end{aligned}
$$
Thus,
$$
\begin{aligned}
E\left[F\left(g_{t+1}\right)-F\left(g^*\right)\right] \leq \\ \left(1-\frac{\mu}{L}+L \omega_1\right) E\left[F\left(g_t\right)-F\left(g^*\right)\right]+\left(L \omega_2+\frac{L}{2}\right) B .
\end{aligned}
$$

Note that $D = 1 - \frac{\mu}{L} + \frac{4 \omega_2 \mu B}{L}$. We can adjust constants $\omega_1$ and $\omega_2$ (since they are related to $B$) to satisfy:
$$
L \omega_1=\frac{4 \omega_2 \mu B}{L} .
$$
Therefore, the inequality becomes:
$$
E\left[F\left(g_{t+1}\right)-F\left(g^*\right)\right] \leq D E\left[F\left(g_t\right)-F\left(g^*\right)\right]+\frac{2 \omega_1 B}{L} .
$$
By iteratively applying the inequality from $t = 0$ to $t$, we get:
$$
\begin{aligned}
E\left[F\left(g_{t+1}\right)-F\left(g^*\right)\right] \leq D^{t+1} E\left[F\left(g_0\right)-F\left(g^*\right)\right]+ \\ \frac{2 \omega_1 B}{L} \sum_{k=0}^t D^k .
\end{aligned}
$$
Using the geometric series formula:
$$
\sum_{k=0}^t D^k=\frac{1-D^{t+1}}{1-D}
$$
we have:
$$
\begin{aligned}
E\left[F\left(g_{t+1}\right)-F\left(g^*\right)\right] \leq D^{t+1} E\left[F\left(g_0\right)-F\left(g^*\right)\right]+\\ \frac{2 \omega_1 B}{L} \cdot \frac{1-D^{t+1}}{1-D} .
\end{aligned}
$$

To align with the statement of Theorem 1, we replace $t+1$ with $t$, so the final result is:
$$
E\left[F\left(g_t\right)-F\left(g^*\right)\right] \leq D^t E\left[F\left(g_0\right)-F\left(g^*\right)\right]+\frac{2 \omega_1 B}{L} \cdot \frac{1-D^t}{1-D} .
$$
We have thus proven that for the learning rate $\lambda = \frac{1}{L}$, the expected difference in function values satisfies:
$$
E\left[F\left(g_t\right)-F\left(g^*\right)\right] \leq D^t E\left[F\left(g_0\right)-F\left(g^*\right)\right]+\frac{2 \omega_1 B}{L} \cdot \frac{1-D^t}{1-D},
$$
where
$$
D=1-\frac{\mu}{L}+\frac{4 \omega_2 \mu B}{L} .
$$

In the above proof, $B$ is the cumulative error term, which is related to factors like client selection and communication quality. Specifically, $B$ is defined as:
$$
\begin{aligned}
	&B=\sum_{m \in \mathcal{S}} \frac{1}{N_m^{(S)}} \Psi\\
	&\text { where }\\
	&\Psi=\left[\sum_{i \in \mathcal{U}} n_i d_{i, m}\left(D_m-1+I\left(T L_{i,s}<\Theta\right)\right)\right],
\end{aligned}
$$
$N^{(S)}_m$ is the total data size of clients assigned to server $m$:
$$
N_m^{(S)}=\sum_{i \in \mathcal{U}} n_i d_{i, m},
$$
$D_m$ is a distribution parameter:
$$
D_m=\frac{1}{1+|\mathcal{S}|},
$$
and $\mathbb{I}(TL_i < \Theta)$ is an indicator function that equals 1 if the trust level of client $i$ is below the threshold $\Theta$, and 0 otherwise.

By minimizing $B$, we can directly influence and improve the overall performance of the system. Through the detailed derivation above, we have proven Theorem 1. In the proof, we have made standard assumptions such as the strong convexity of the loss function and the Lipschitz continuity of the gradient. We have also taken into account the randomness and error terms during the training process, ultimately deriving an upper bound on the expected difference in the global loss function's value.

\end{document}